\documentclass[conference]{IEEEtran}
\usepackage{times}
\usepackage{times}
\usepackage{epsfig}
\usepackage{graphicx}
\usepackage{amsmath}
\usepackage{verbatim}
\usepackage{amssymb}
\usepackage{CJKutf8}
\usepackage{caption}

\usepackage{mathrsfs}
\usepackage[numbers]{natbib}
\usepackage{multicol}
\usepackage[bookmarks=true]{hyperref}

\begin{document}

\title{Multi-Granularity Modularized Network for Abstract Visual Reasoning}


\author{\authorblockN{ Xiangru Tang$^{1}$,       
   Haoyuan Wang$^{2}$,       
   Xiang Pan$^{3}$,       
   Jiyang Qi$^{2}$
  }
\authorblockA{
  $^{1}$University of the Chinese Academy of Sciences, China\\
  $^{2}$Huazhong University of Science and Technology, China\\
  $^{3}$National University of Singapore, Singapore\\
  xarutang@gmail.com
}
}


%

\maketitle

\begin{abstract}
Abstract visual reasoning connects mental abilities to the physical world, which is a crucial factor in cognitive development. 
Most toddlers display sensitivity to this skill, but it's not easy for machines. Aimed at it, we focus on Raven's Progressive Matrices Test, designed to measure cognitive reasoning~\cite{raven2000raven}.
Recent work designed some black boxes to solve it in an end-to-end fashion, but they're incredibly complicated and difficult to explain. Inspired by cognitive studies, we propose a Multi-Granularity Modularized Network (\textup{MMoN}) to bridge the gap between the processing of raw sensory information and symbolic reasoning. Specifically, it learns modularized reasoning functions to model the semantic rule from the visual grounding in a neuro-symbolic and semi-supervision way.  To comprehensively evaluate \textup{MMoN}, our experiments are conducted on the dataset of both seen and unseen reasoning rules. The result shows that \textup{MMoN} is well suited for abstract visual reasoning and also explainable on the generalization test.

\end{abstract}

\section{Introduction}

Can an agent do relational and analogical visual reasoning as well as a toddler?
Moreover, can an agent solve reasoning tasks it has never seen before? 

Abstract visual reasoning is a remarkable cognitive mechanism for humans to achieve logical conclusions in the absence of physical objects, specific instances, or concrete phenomena.
And here, the capacity of reasoning is a generalization about relations and attributes primarily, instead of concrete objects.
More importantly, current machine learning techniques are data-hungry and brittle—they can only make sense of patterns they've seen before. Using current methods, an algorithm can gain new skills by exposure to large amounts of data, but cognitive abilities that could broadly generalize to many tasks remain elusive. Thus, there's a question about what happens if the agent meets a new and unseen reasoning type. And also, we want to know what is the thinking in images and abstract reasoning for machines. 

To deal with these issues, we focus on abstract visual reasoning, offer the potential for more human-like abstraction and reasoning. Correctly, we verify our agent on Raven's Progressive Matrices (RPM) Test, designed to measure abstract visual reasoning. It's also used to test the human's capacity of non-verbal cognitive functioning in some public exams.
In the measurement, the agent is showed with a $3\times3$ matrixes with geometric designs. Given eight candidates of the missing layout, the agent is aimed at choosing the correct layout, and need to follow the analogical relations' rule and figure out the specific pattern in this matrix ~\cite{zhang2019raven}, based on the Spearman's two-factor theory of intelligence~\cite{weiten2007psychology}.

Unlike existing work in measuring abstract visual reasoning using  RPM~\cite{santoro2018measuring}, we simulate and validate our designs on RAVEN~\cite{zhang2019raven} test because RAVEN test establishes a semantic link between vision and thinking by providing tree-based structure representation. Previous work~\cite{zhang2019raven,zheng2019abstract, zhang2019learning, wang2020abstract} design extremely complex models to do representation and reasoning in an end-to-end fashion. But their models are tedious and, therefore, hard to explain, besides, the structure information is not well utilized. Most importantly, they cannot easily extrapolate their knowledge to new situations.

Hoping to understand better how machines understand this task, we aim to figure out whether the computer can learn the rule (semantic) from the visual sensory information. Ask for toddlers, and toddlers must rely on intrinsic cognitive functions for logical conclusions.
Inspired by these cognitive studies, we equip our model with simple modularized reasoning functions that is jointly trained with the perception backbone in a neuro-symbolic way.  Toddlers can be attuned to relationships between features of objects, actions, and the physical environment. We adopt module network\cite{andreas2016neural}, and each module is for each rule in our case. To train it, we want to take our rules as the target of a latent semantic parser. And the goal is to recover that. Meta target information is then utilized to restrict the space of potential semantic parser that we consider, which provides a certain level of intelligence. To determine our model's efficiency, we verify our model on the RAVEN dataset compared to various baselines. Furthermore, we design four generalization test to demonstrate the improved ability to deal with unseen reasoning rules.



\section{Related Work}

Raven’s Progress Matrices problem is widely used to test the capability
of abstract reasoning. In recent years, different models and datasets are designed to lift the reasoning ability of modern vision systems.
Inspired by RPM, \cite{santoro2018measuring} built the first large-scale RPM dataset named PGM, and proposed a relational model Wild Relation Network (WReN) leverage representation of pair-wise relations for each choice.
Then, \cite{steenbrugge2018improving} made use of pre-trained Variational Auto Encoder to improve the generalization performance of WReN\cite{santoro2018measuring}.
\cite{zhang2019raven} generated a new RPM-style dataset RAVEN with structured representation and proposed Dynamic Residual Tree (DRT), which considers annotations of image structure. 
Both PGM and RAVEN are designed to be easy to recognize but hard to reason.
\cite{zheng2019abstract} proposed a student-teacher architecture to deal with distracting features.
More recently, \cite{wang2020abstract} used a multi-layer multiplex graph to capture multiple relations between objects.
Besides, \cite{zhuo2020solving} modified ResNet\cite{he2016resnet} to reduce overfitting, and proposed MCPT to solve RPM problems in an unsupervised manner.

\begin{figure*}[t]

\centering
\includegraphics[scale=0.45]{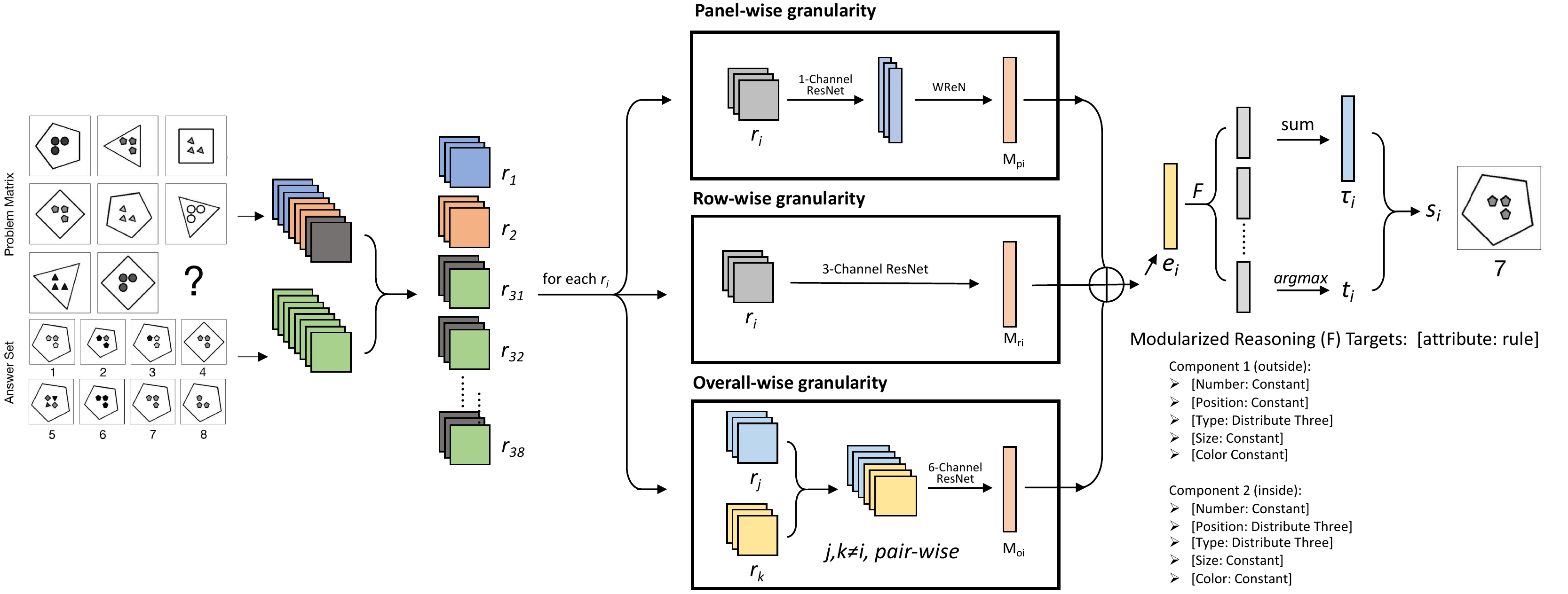}
\caption{Multi-Granularity Modularized Network}
\label{model}
\end{figure*}

Many previous studies have utilized modular neural architectures for various tasks. Such as \cite{andreas2016nmn} assembled networks flexibly from a collection of specialized substructures to answer questions. And \cite{mao2018the} could learn a good representation of visual concepts and semantic parsing of sentences from images and question-answer pairs jointly, even without explicit supervision because it utilizes the different modules to extract different information.



\section{Approach}

\subsection{Problem Formulation}

The task is designed to measure non-verbal, cognitive, and abstract reasoning. In the task's setting, the agent is showed with a $3\times3$ matrixes with geometric designs. And most importantly, the last diagram is missing. Given eight candidates of the missing layout, the agent is aimed at choosing the correct layout, and need to follow the analogical relations' rule. See the example in Figure \ref{model}, specifically in this problem, it is an inside-outside structure
in which the external component is a layout with a single centered object, 
and the inside element is a $2\times2$ grid layout. 
The rules are listed in Figure~\ref{model}.
The compositional nature of the
rules makes this problem a difficult one, and the correct answer is seven.

The task could be formally defined as: Given $N$ training samples, denoted as $\{\{ x_i, y_i, m_i\}\}_{i=1}^N$, where $x_i$ is the input images contains 8 content panels $P = \{p_{11}, p_{12} ... p_{33}\}$ and 8 candidate answers $A = \{a_1, a_2, ...a_8\}$. $y_i$ is the label and $m_i$ is the meta target of training sample. Meta-target is a tensor containing attributes and rules of $x_i$. The input sample has a rule sets $R = \{R_1, R_2...R_n\}$, where $R_i$ is a tuple containing two elements: $R_i = (\alpha, t)$, which means that for a certain row or column in $x_i$, the attribute $\alpha$, has a rule $t$. Suppose $\alpha$ was sampled from a attribute set $\{\alpha_1, \alpha_2, ...\alpha_p \}$. The input of the model input images $x_i$ and meta-target $m_i$ given by the dataset. In detail, meta-target is a multi-hot vector consists of attribute part (to represent $\alpha$) and rule part (to represent $t$): each position of vector represents an attribute or rule, 1 for existing, 0 for not existing. This information is used for learning rules of training samples.

Specifically, there are 4 types of rules in our setting: Constant, Progression, Arithmetic,
and Distribute Three. As shown in Figure 1, they are denoted as [attribute: rule] pair.

\subsection{Multi-Granularity Modularized Network}
There are two vector spaces in our architecture, the scope of visual representation, and the scope of reasoning rules. Given modules corresponding to different attribute types, we need to learn a projection from high-dimensional representation space to low-dimensional rule space and find the most similar embedding.

\subsubsection{Multi-Granularity Sensory Representation }

Inspired by principles of psychological development, 
The capacity for human abstract visual reasoning develops from the initial reasoning about physical objects, especially some concrete objects. Also, this capacity then develops from the subsequent formation of categories and schemas~\cite{luszcs1989psychological}. Inspired by this hierarchical reasoning strategies, we incorporate three-granularity hierarchical features from three levels of granularity: panel-level $M_p$, row-level $M_r$, and overall-level $M_o$. This multi-granularity sensory representation captures both coarse-grained and fine-grained features effectively. Also, the representation of each panel is coupled and
interacts with each other.

\begin{table*}[t]
\centering
\begin{tabular}{|c|c|c|c|c|c|c|c|c|}
\hline
models &Avg
 &  Center
 &  2*2Grid
 &  3*3Grid
 &  L-R 
 &  U-D
 &  O-IC
 &  O-IG \\
\hline
Random
 &  12.50\%  & 12.50\%   &  12.50\%  &  12.50\%  &  12.50\%  &  12.50\% & 12.50\% & 12.50\% \\
LSTM\cite{zhang2019raven}
 & 13.07\% & 13.19\%  & 14.13\% & 13.69\% & 12.84\% & 12.35\% & 12.15\% & 12.99\%\\
 LSTM+DRT\cite{zhang2019raven}
 & 13.96\% & 14.29\%  & 15.08\% & 14.09\% & 13.79\% & 13.24\% & 13.99\% & 13.29\%\\
 
 CNN\cite{zhang2019raven} & 36.97\% & 33.58\%  & 30.30\%  & 33.53\% & 39.43\% & 41.26\% & 43.20\% & 37.54\%\\
CNN+DRT\cite{zhang2019raven} & 39.42\% & 37.30  & 30.06  & 34.57 & 45.49 & 45.54 & 45.93 & 37.54\\
\hline

ResNet-18+MLP+DRT\cite{zhang2019raven}
 & 59.56\% & 58.08\% & 46.53\% & 50.40\% & 65.82\% & 67.11\% & 69.09\% & 60.11\% \\
 
 ResNet-50+MLP+DRT\cite{zhuo2020solving}
 & 86.26\% & 89.45\% & 66.60\% & 67.95\% & 97.85\% & 98.15\% & 96.60\% & 87.20\%\\
 \hline
 ~
WReN\cite{santoro2018measuring}  & 14.69\% & 13.09\% & 28.62\% & 28.27\% & 7.49\% & 6.34\% & 8.38\% & 10.56\%\\

LEN\cite{zheng2019abstract}  & 72.9\% & 80.2\% & 57.5\% & 62.1\% & 73.5\% & 81.2\% & 84.4\% & 71.5\%\\
LEN + Teacher Model\cite{zheng2019abstract}   & 78.3\% & 82.3\% & 58.5\% & 64.3\% & 87.0\% & 85.5\% & 88.9\% & 81.9\%\\
MXGNet\cite{wang2020abstract} & 83.91\% &/  & / &/& /& /& / & /\\
 \hline
 
\textup{MMoN}& 83.01\% & 92.06\% & 82.84\% & 63.44\% & 82.33\% & 78.29\% & 80.45\% & 82.54\%\\

\textup{MMoN}(meta-target)& 87.04\%& 95.12\%  & 90.14\% &78.82\%& 89.45\%& 84.72\%& 82.68\% & 88.27\%\\
\hline
\end{tabular}
\caption{Testing accuracy of different models on RAVEN.}
\label{tab1}
\end{table*}

\begin{table*}[htbp]
\centering
\begin{tabular}{|c|c|c|c|c|c|c|}
\hline
models 
&  Center
 & L-R 
 &U-D
 & O-IC

 &  2*2Grid
 &  3*3Grid \\
\hline
Random
 &  12.50  & 12.50   &  12.50  &  12.50  &  12.50  &  12.50  \\
ResNet-18+MLP+DRT\cite{zhang2019raven}
  & 51.87 & 40.03& 35.46 & 38.84&  38.69 & 39.14\\
 
 ResNet-50+MLP+DRT\cite{zhuo2020solving}
 & 60.80  & 43.65&  41.40 & 43.65& 42.24 & 43.87\\

 \hline
 
\textup{MMoN}& 59.47& 40.28  & 38.91& 41.11& 39.84 & 42.55\\

\textup{MMoN}(meta-target)& 62.49 &45.21& 43.68& 44.56& 45.16 & 47.25\\
\hline
\end{tabular}
\caption{Generalization test. First, the model is trained on Center and
tested on three other figure configurations, and then 3*3Grid column means the model is trained on 2*2Grid and tested on 3*3Grid, 2*2Grid column implies the model is trained on 2*2Grid and tested on 3*3Grid.
}
\label{tab2}
\end{table*}

\begin{table*}[htbp]
\centering
\begin{tabular}{|c|c|c|c|c|c|c|c|}
\hline

models 
&  Center
 & L-R 
 &U-D
 & O-IC(single)
 & O-IC(four)
 &  2*2Grid
 &  3*3Grid \\
\hline
Random
 &  12.5000  & 12.5000   &  12.5000  &  12.5000  &  12.5000  &  12.5000   &  12.5000\\
LSTM\cite{zhang2019raven}
 & 12.0192 & 13.2212 & 11.7788 & 12.0192& 10.0962 & 13.4615 & 11.7788\\
 LSTM+DRT\cite{zhang2019raven}

 & 12.2596 &12.2596& 9.8558 & 12.5000& 13.7019& 12.5000&10.0962\\
 
 CNN\cite{zhang2019raven} & 12.0192 & 12.0192 & 11.0577 & 13.9423 &  10.5769 & 10.3365 & 12.5000 \\
CNN+DRT\cite{zhang2019raven} 
 & 11.5385  & 12.2596 & 14.9038 & 12.5000&11.0577&13.4615& 10.8173\\

\hline

ResNet-18+MLP+DRT\cite{zhang2019raven}
 & 19.9519 & 17.3077& 20.1923 & 14.6635&  17.7885&18.2692 & 15.8654\\
ResNet-50+MLP+DRT\cite{zhang2019raven}
   & 31.4904 & 40.8654 & 37.0192 & 37.9808 & 31.7308& 30.2885 & 25.7212\\
 
 \hline

\textup{MMoN}& 33.9500 & 38.5000  & 17.7500& 39.4499& 31.3000 & 35.8500& 38.5499\\

\textup{MMoN}(meta-target)& 40.1534 &43.7400& 22.1868& 41.8625& 37.4919 & 43.9735&47.1300\\
\hline
\end{tabular}
\caption{Generalization test. The model is trained on dataset without rule of disturbted three and
tested on rule of disturbted three. And then another model is trained on dataset without progression but
tested on rule of progression.}
\label{tab3}
\end{table*}

\textbf{Panel-wise granularity ($\mathbf{M_p}$):}
lt takes each panel as input and handles the attributes
of inside graphical element. Moreover, we take
the correlations among panels of the same row into consideration, and apply Relation Network~\cite{santoro2017simple}  to obtain this inner relationship. For each panel $p_{i1}, p_{i2}, p_{i3}$ in row $r_i$, firstly we use Residual Network\cite{he2016resnet} to extract the features ($\mathfrak{F}$) of each: $\mathfrak{R}_{ij} = \textsl{ResNet}(p_{ij})$. Then WReN is used to extract the representation of pair-wise relationship of 3 panels in a row $r_i$: 

\begin{equation}
\textrm{M}_{pi} = \textsl{WReN}([\mathfrak{R}_{i1}, \mathfrak{R}_{i2}, \mathfrak{R}_{i3}])
\end{equation}

\textbf{Row-wise granularity ($\mathbf{M_r}$):} 
Furthermore, the network of individual hierarchy takes each row as input. In Raven, the same rules are applied to rows. Motivated by this, we stack the three panels in row $r_i$ together (instead of treating each panel separately as we did in the previous section). Then it was fed to a pre0trained 3-channel ResNet to encode the entire row with a compact embedding.
\begin{equation}
\textrm{M}_{ri} = \textsl{ResNet}([p_{i1}, p_{i2}, p_{i2}])
\end{equation}

\textbf{Overall-wise granularity ($\mathbf{M_o}$):}
Considering rules of the third row are the same as rules of $r_1, r_2$, it's essential to take the two rows together as input and
jointly learns the rule patterns underlying the two rows.  Thus, we perform pair-wise embedding to capture the interaction between two rows: $(r_1, r_2), (r_1, r_{31}), (r_1, r_{32})...(r_1, r_{38})$ and $(r_2, r_{31}), (r_2, r_{32})...(r_2, r_{38})$ just like what WReN does.
 $c_{ij}$ combines $r_i, r_j$, is obtained and passed to $M_o$ as input. $M_o$ treat each combination $c_{ij}$ as a whole and uses a 6-channel ResNet to take the pair-wise relationships among rows into consideration.
\begin{equation}
\textrm{M}_{oi} = \sum_{\substack{j, k\neq i\\ j, k=1 }}^3 \textsl{ResNet}(c_{jk})
\end{equation}

\subsubsection{Modularized Reasoning}
Here, we need to inference the corresponding rules from the representation and then obtain the answer. There are learning objects, and a bunch of modularized functions corresponds to different rules. So each module presents a specific rule, and we regard each rule is an operation.  Then, given the attribute and the rule, we use a simple MLP ($\mathfrak{f}$) to tell us how correct the candidate is. 
The goal of functions ($\mathfrak{f}$) is to learn the right parameterization of modules to gain the right rule, like the size changes or the color remains. 
In this neuro-symbolic way~\cite{mao2019neuro,han2019visual}), 
the signals for learning modules come from sensory representation, and the final candidate selection could jointly train them. We can end up with a rule embedding that is closed to the projection.
For the training part,
we can start from randomly initialized modules. Moreover, we feed the meta-target as semi-supervised signals.

To bind an attribute to a specific $f_i$, we used the meta-target information in the dataset as supervision. The meta-target records the attributes and rules included in the current training sample. During training, we only train the network corresponding to the present sample attributes. For example, the attributes of the current sample are $\alpha_1, \alpha_3, \alpha_5$, then we only train networks $f_1, f_3, f_5$ and freeze other MLP. In this way, the MLP $f_i$ was bind to a unique attribute of $\alpha_i$.


Moreover, we can define the proposed modularized functions as $F=\{f_1, f_2, f_3....f_p\}$, corresponding to the attribute set $\{\alpha_1, \alpha_2, ...\alpha_p\}$).  For each attribute, the multi-granularity representation, denoted as  $e_i = \textrm{M}_{pi} + \textrm{M}_{ri} + \textrm{M}_{oi}$, are fed to $\mathfrak{F}$ to obtain the transformation $t$ of one rule $r$. 

\begin{equation}
    \tau_{i} = \sum_{j=1}^p(f_j(e_i))
\end{equation}

Then modularized functions ($\mathfrak{f}$) are used to inference the transformation $t$ on a specific attribute with meta-targets. Cosine similarity is applied to inference the  $t$, which most similar to correct $t^*$, given by meta-target. Concretely we compute similarity between $\tau_3$ and $\tau_1, \tau_2$ to choose the best candidate.

\begin{equation}
    t_i = argmax(\mathfrak{f}_s(\mathfrak{f}_j(e_i), t^*))
\end{equation}

\begin{equation}
    s_i =  \sum_{j=1}^p \{\mathfrak{f}_s(\tau_{3}, \frac{\tau_1 + \tau_2}{2})  + \mathfrak{f}_s(t_i, t^*)\}
\end{equation}

As for the inference part, we apply the cosine similarity function as a score function $f_s$ to calculate the score of $R_3$. The scores in the three granularities are denoted as $s_p, s_r, s_o$, respectively. Finally, We choose the one with the highest total score in $R_3$, and the corresponding candidate answer is the output of the model.

Our model finally choose
one image from the candidate set to complete the matrix correctly, namely satisfying the underlying
rules in the matrix.
\begin{equation}
output=argmax(\{s_1, s_2...s_8\})
\end{equation}

\begin{equation}
\textsl{Loss}(s, y) = \textsl{CE}(s, y) + \lambda (s_m - \sum_{\substack{i=1 \\ i\neq m}}^8 s_i)
\end{equation}

where  $y$ is the target label, and $a_m$ is the correct answer. $\lambda$ is an adjustable hyperparameter, controlling the weight of these two losses. The weak supervision comes from concept space and answer. The meta target gives a constraint on what kind of operations we are going to use, and the answer gives us the constraints that what output what each of the functions ($\mathfrak{f}$) it should be.







\section{Experiment}
In our experiment, the dataset is split into training, validation, and testing with the ratio 6:2:2, respectively. 
The input images of the model are resized to $80 \times 80$. In the training step, because the input channel of 3 levels is different, we modify the first channel of each ResNet to 1, 3, 6, respectively, fitting the shape of the input. In detail, our model contains MLP with the same structure(with linear and dropout layers).  In Raven, there are two components for each sample, and each component has five attributes. And we use Adam as an optimizer for training and set $\beta$ of Adam to $(0.9, 0.999)$. Our code is available at https://github.com/creeper121386/RAVEN-test.

\begin{figure}[htbp]
\centering
\includegraphics[scale=0.22]{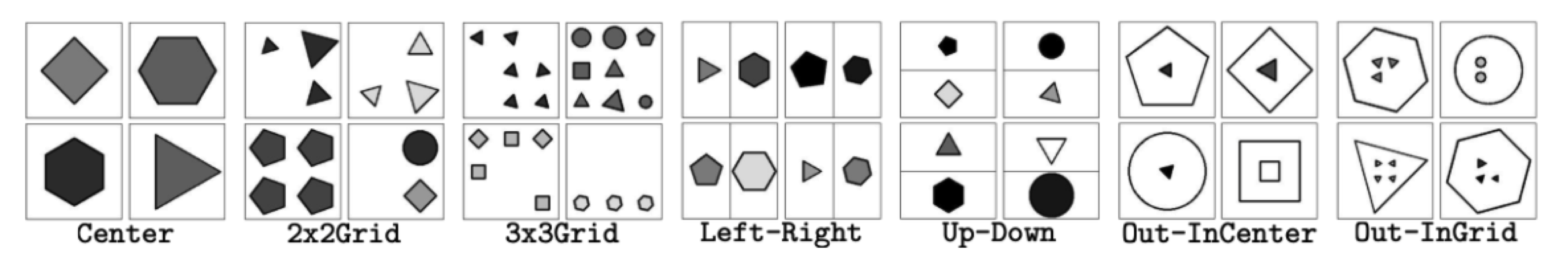}
\caption{Different kinds of  components in the dataset.}
\label{show}

\end{figure}

In total, we have seven configurations, as shown in Figure \ref{show}. To test the generalization performance of models on unseen data, we design two experiments: (a) the models are trained on a subset of data but tested on another subset with unseen layout, (b) the models are trained on a subset of data without one specific rule, and then tested on another subset with this rule. Results can be found in Table \ref{tab2} and \ref{tab3}. The result shows that \textup{MMoN} is well suited for abstract visual reasoning (see Table \ref{tab1}) and also explainable on the generalization.

\section{Conclusion}

We propose a novel Multi-Granularity Modularized Network, which performs high accuracy and maintains the stability of the model on different layouts of Raven.

\section{Acknowledgments}
We would sincerely thank Rodolfo Corona for useful discussions and assistance with data analysis. This work is substantially supported by by the National Natural Science Foundation of China under the grant number 61572223 and the University-Industry Collaborative Education Program between the Ministry of Education of China and Google Information Technology (China) Co., Ltd. (PJ190496).

\begin{CJK}{UTF8}{gbsn}

\end{CJK}


\bibliographystyle{plainnat}
\bibliography{references}
\end{document}